\documentclass[10pt,twocolumn,letterpaper]{article}

\usepackage{multirow}
\usepackage{booktabs}
\usepackage{array}
\usepackage{geometry}
\usepackage{caption}
\usepackage{amsmath}

\usepackage{amssymb} 

\usepackage[pagenumbers]{cvpr} 

\usepackage{float}
\usepackage{newtxtt}
\usepackage{tikz} 
\usepackage{listings}
\usepackage[table]{xcolor}
\usepackage{booktabs}
\usepackage{tabularx}
\usepackage{multirow}
\usetikzlibrary{shapes.geometric, arrows.meta, positioning, calc}

\definecolor{codegreen}{rgb}{0,0.6,0}
\definecolor{codegray}{rgb}{0.5,0.5,0.5}
\definecolor{codepurple}{rgb}{0.58,0,0.82}
\definecolor{codeblue}{rgb}{0.13,0.13,0.82}
\definecolor{backcolour}{rgb}{0.95,0.95,0.92}

\lstdefinelanguage{OpenSCENARIO}[]{Python}{
    morekeywords={use, scenario, map, with, keep, environment, vehicle, stationary_object, var, speed, length, do, serial, parallel, one_of, wait, emit, fall, rise, elapsed},
    alsoletter={@}, 
}

\definecolor{cvprblue}{rgb}{0.21,0.49,0.74}
\usepackage[pagebackref,breaklinks,colorlinks,allcolors=cvprblue]{hyperref}

\def\paperID{*****} 
\def\confName{CVPR}
\def\confYear{2026}

\title{Compiling OpenSCENARIO 2.1 for Scenario-Based Testing in CARLA}

\author{Thoshitha Gamage\\
Department of Computer Science\\
Southern Illinois University Edwardsville\\
Edwardsville, IL\\
{\tt\small tgamage@siue.edu}
\and
Lasanthi Gamage\\
Department of Computer Science\\
Webster University\\
Webster Groves, MO\\
{\tt\small lasanthigamage67@webster.edu }
}

\begin{document}

\maketitle
\begin{abstract}
While the ASAM OpenSCENARIO 2.1 Domain-Specific Language (DSL) enables declarative, intent-driven authoring for Scenario-Based Testing (SBT), its integration into open-source simulators like CARLA remains limited by legacy parsers. We propose a multi-pass modern compiler architecture that translates the OpenSCENARIO 2.1 DSL directly into executable CARLA behaviors. The pipeline features an ANTLR4 frontend for Abstract Syntax Tree (AST) generation, a semantic middle-end, and a runtime backend that synthesizes deterministic \texttt{py\_trees} behavior trees. Mapping the standardized domain ontology directly to CARLA's procedural API via a custom method registry eliminates the need for external logic solvers. A demonstrative multi-actor cut-in and evasive maneuver, selected from a wider suite of validated scenarios, confirms the compiler's ability to process concurrent actions, dynamic mathematical expressions, and asynchronous signaling. This framework establishes a functional baseline for reproducible, large-scale SBT, paving the way for future C++ optimizations to mitigate current Python-based computational overhead.
\end{abstract}    
\section{Introduction}
\label{sec:intro}

The transition toward highly automated and autonomous vehicles necessitates a paradigm shift in verification and validation methodologies. Relying on traditional, distance-based field tests to uncover safety-critical edge cases is economically infeasible and statistically insufficient~\cite{kalra2016driving}. Consequently, the industry has widely adopted Scenario-Based Testing (SBT), a paradigm that accelerates evaluation by systematically isolating and simulating complex traffic interactions within high-fidelity virtual simulation environments like CARLA~\cite{dosovitskiy2017carla}.

SBT demands a standardized, machine-readable language capable of describing these interactions with both precision and flexibility. Historically, the industry relied on ASAM's (Association for Standardization of Automation and Measuring Systems)  \textit{de facto} standard for interoperability: ASAM OpenSCENARIO XML 1.x~\cite{asam2020openscenarioxml}. While this XML schema is optimized for defining predictable trajectory playback, it forces a trade-off between expressive flexibility and maintainability as scenario complexity grows. To address this, ASAM bifurcated the standard by releasing ASAM OpenSCENARIO Domain Specific Language (DSL) standard. The first version (v2.0) was released in July 2020 and a more mature and a comprehensive version (v2.1) was released in March 2024 \cite{asam2024openscenariodsl}\footnote{Since finalizing OpenSCENARIO v2.1 standard, ASAM continues to refine the DSL, currently fielding public reviews for the 2.2.0 release candidate as of early 2026. However, v2.1 remains the foundational finalized library supported by this architecture.}. The DSL standard introduces a human-readable, declarative syntax that enables intent-driven scenario authoring. It allows researchers to express complex maneuvers as reusable, parameterized logic across abstract, logical, and concrete layers, significantly reducing boilerplate. Table~\ref{tab:osc_comparison_1_3} provides a comparison between these two standards to highlight this point. 

\begin{table}[!hbt]
\centering
\small
\rowcolors{2}{gray!10}{white}
\caption{Comparison of OpenSCENARIO v1.3 (XML) and v2.1 (DSL)}
\label{tab:osc_comparison_1_3}
\begin{tabularx}{\columnwidth}{@{} 
    >{\raggedright\arraybackslash}p{0.22\columnwidth} 
    >{\raggedright\arraybackslash\hsize=0.85\hsize}X 
    >{\raggedright\arraybackslash\hsize=1.15\hsize}X 
    @{}}
\toprule
\rowcolor{white} \textbf{Feature} & \textbf{OSC 1.3 (XML)} & \textbf{OSC 2.1 (DSL)} \\
\midrule
\textbf{Format} & Schema-based (XML) & Text-based DSL (EBNF) \\
\textbf{Hierarchy} & Storyboard  $\rightarrow$ Act  $\rightarrow$ Maneuver  $\rightarrow$ Event  $\rightarrow$ Action & Declarative, intent-driven structure \\
\textbf{Logic} & Trigger-based conditions & Full expressions, conditions, and logic \\
\textbf{Param\-eter\-ization} & Parameters and catalogs & Native variables and reusable abstractions \\
\textbf{Modularity} & Limited reuse via catalogs & High modularity and composability \\
\textbf{Behavior Modeling} & Explicit action definitions & Actions + Modifiers + Conditions separation \\
\textbf{Semantic Validation} & Syntax (XSD) only & Supports ontology-based semantic validation \\
\textbf{Readability} & Verbose ($\sim$150--300 lines) & Concise ($\sim$20--50 lines) \\
\textbf{Integration} & Widely supported in tools & Requires DSL parser/compiler, limited support \\
\textbf{Execution Model} & Event-driven storyboard execution & Intent-driven, composable execution model \\
\textbf{Extensibility} & Limited & High (DSL + custom constructs) \\
\bottomrule
\end{tabularx}
\end{table}

Integrating the OpenSCENARIO DSL into established open-source engines, nonetheless, has revealed a substantial implementation gap. Widely used orchestration frameworks, such as \texttt{ScenarioRunner}~\cite{carlateam2024scenariorunner}, which operate on top of simulation engines like CARLA, historically provided only partial support. Furthermore, most existing implementations lack a robust compiler for the finalized OpenSCENARIO v2.1 specification. Within these frameworks, the legacy interpreters utilize brittle parsing architectures that trigger syntax errors when attempting to ingest the bifurcated, finalized standard libraries (\texttt{types.osc} and \texttt{domain.osc}). The net result is the inability to leverage the full expressive power of the finalized DSL for research and industrial-grade scenario validation.

This paper addresses this bottleneck by presenting a fully compliant  OpenSCENARIO v2.1 compiler architecture integrated directly with the CARLA/\texttt{ScenarioRunner} environment. Our work utilizes ANTLR4~\cite{parr2013antlr4} to implement an EBNF-driven compiler. This multi-stage system generates an Abstract Syntax Tree (AST) that ensures type safety and semantic validity, ultimately mapping domain modifiers to CARLA’s atomic behaviors through deterministic Behavior Trees (\texttt{py\_trees}). The following sections detail a three-stage compiler architecture and outline the simulation ontology features supported by the implementation. Furthermore, a complete scenario file is provided to demonstrate a functional example of these features in practice.
\section{Related Work} \label{sec:rel_work}

\begin{figure*}[!ht]
    \centering
    \includegraphics[width=\textwidth]{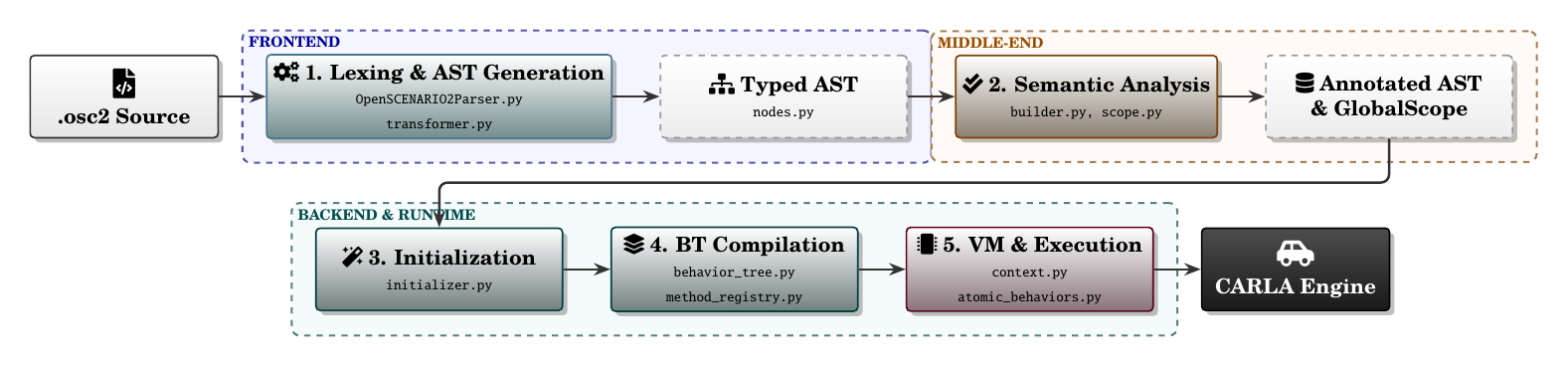}
    \caption{The 3-stage OSC2 Execution Pipeline mapping DSL definitions to CARLA behaviors.}
    \label{fig:pipeline}
\end{figure*}

The OpenSCENARIO DSL formalization joins an expanding body of research into domain-specific languages for autonomous driving. It complements existing frameworks like the probabilistic Scenic \cite{fremont2019scenic}, the map-centric GeoScenario \cite{queiroz2019geoscenario}, and testing-focused DSLs such as Paracosm \cite{majumdar2019paracosm}. These approaches offer complementary strengths, including stochastic scenario generation, high-fidelity geographic modeling, and parameterized test case design. However, they are typically developed as standalone frameworks and often lack standardized semantics, limiting interoperability across tools and simulation platforms.

In contrast, the OpenSCENARIO DSL aim to unify scenario representation through a declarative DSL that integrates \textit{spatial}, \textit{temporal}, and \textit{behavioral} semantics within a single framework. The OpenSCENARIO v2.1 standardizes many of these concepts, introducing composition operators for orchestrating complex multi-actor interactions and enabling the specification of Key Performance Indicators (KPIs) directly within scenario definitions \cite{asam2024openscenariodsl}. Furthermore, it is designed to interoperate with complementary standards in the ASAM OpenX ecosystem, such as OpenDRIVE and OpenLABEL, and can be integrated with simulation ontologies to support semantic reasoning about physical feasibility and domain-specific constraints.

Parsing OpenSCENARIO DSL grammar into an Abstract Syntax Tree (AST) relies heavily on modern ANTLR4-based lexical analyzers, most notably the community-driven \textit{py-osc2} framework \cite{pmsf2024pyosc2}. Once parsed, translating this declarative logic into continuous simulation execution is a critical bottleneck. RoadLogic \cite{bartocci2026roadlogic} proposes translating the OpenSCENARIO DSL into a symbolic automaton representation, utilizing Answer Set Programming (ASP) to solve high-level planning constraints before passing them to a motion planner. While logically rigorous, this approach introduces the computational overhead of an external logic solver. 

Similarly, the Yase framework \cite{kaefer2023yase} explores a multi-stage compiler architecture that maps an AST to a behavior tree, but functions primarily as an agnostic middle-end integrated into the openPASS ecosystem. Other recent frameworks, such as VIVAS \cite{goyal2025vivas} and BeSimulator \cite{wang2024besimulator}, have demonstrated the viability of utilizing \textit{py\_trees} for managing complex agent perception and action states dynamically in simulators like CARLA. In contrast to architectures requiring intermediate external solvers, our approach directly compiles generated AST into execution-ready \textit{py\_trees} natively within CARLA, allowing for real-time, tick-by-tick simulation by mapping the \texttt{domain.osc} actions to CARLA's localized Python API.

Recent literature also highlights the use of Large Language Models (LLMs) to automatically generate DSL scripts from natural language. Frameworks such as Text2Scenario \cite{cai2025text2scenario} utilize models like GPT-4 to parse scenario descriptions and output OpenSCENARIO DSL scripts intended for simulators like CARLA. Similarly, platforms like Chat2Scenario \cite{chat2scenario2024} and top-down generation pipelines such as LeGEND \cite{tang2024legend} demonstrate the capability of LLMs to transform unstructured traffic data, crash reports, and abstract functional scenarios into formal logical parameters. However, these generative approaches focus entirely on the frontend creation of the scenario text, fundamentally assuming a functioning backend compiler exists to execute the output script. Our architecture provides the underlying execution engine required to make such LLM-driven generation pipelines practically viable.

\section{New Compiler Architecture} \label{sec:compiler_architecture}

Despite aforementioned advances, practical adoption of DSLs remains limited by a lack of robust tooling for formal parsing, semantic validation, and simulation integration. To bridge this gap, we introduce a multi-pass compiler architecture that translates OpenSCENARIO v2.1 \textit{(hereafter OSC2)} into executable CARLA behaviors, providing a seamless workflow from standard-compliant authoring to simulation. As illustrated in Figure \ref{fig:pipeline}, the pipeline is structured around three primary compilation stages -- Frontend Parsing, Semantic Analysis, and Backend Execution Generation -- supported dynamically by a Runtime Context Manager. This decoupled approach ensures strict compliance with the OSC2 domain model while allowing flexible mapping to the underlying physics and navigation engines.

\subsection{Lexing and AST Generation}
The first stage of the compilation pipeline is the Frontend, responsible for lexical and syntax analysis. ANTLR4 is utilized to process the raw OSC2 source code against the standard Extended Backus-Naur Form (EBNF) grammar, producing a raw parse tree. 

To map the concrete syntax to the internal logic of the compiler, an \texttt{ASTTransformer} applies the Visitor pattern to the parse tree, generating a typed Abstract Syntax Tree (AST). The AST is implemented utilizing strongly-typed data structures to ensure compile-time type safety. At this stage, the transformation is purely syntactic; no scoping, type checking, or symbol resolution is performed. This guarantees that the AST accurately represents the syntactic structure of the source file independently of its semantic validity.

\subsection{Semantic Analysis and Symbol Resolution}
Once generated, the syntactic AST is passed to the \texttt{ModelBuilder} for Semantic Analysis. Because the OSC2 language supports hierarchical ontological structures---such as actor inheritance (e.g., \texttt{vehicle} inheriting from \texttt{traffic\_participant}), physical type definitions, and forward declarations---the semantic analyzer operates via a two-pass methodology:
\begin{enumerate}
    \item \textbf{Definition Pass:} The compiler traverses the AST to populate the Symbol Table. It constructs a global scope alongside nested lexical scopes for namespaces, structures, and actors. All named entities (variables, methods, actions, and modifiers) are registered as abstract symbols.
    \item \textbf{Resolution Pass:} The compiler traverses the AST a second time to bind symbols to their corresponding type definitions, resolve inheritance chains, and validate constraints. During this phase, implicit fallbacks to standard library domains (e.g., the \texttt{stdtypes} namespace from \texttt{types.osc}) are safely resolved.
\end{enumerate}
Upon successful completion of these passes, the AST is considered semantically valid, annotated with scope contexts, and ready for runtime evaluation.

\subsection{Initialization, BT Compilation, and Execution}
The final stage of the pipeline translates the semantically validated AST into an executable format. Rather than generating static bytecode, the compiler backend synthesizes a dynamic Behavior Tree (BT) utilizing the \texttt{py\_trees} framework. This generation is orchestrated through two primary components:
\begin{itemize}
    \item \textbf{Behavior Tree Builder:} This module acts as the code generator, mapping OSC2 control flow directives (\texttt{do}, \texttt{serial}, \texttt{parallel}, \texttt{one\_of}) into corresponding Behavior Tree composites. It wraps durative actions with timeout constraints and maps OSC2 event triggers to native condition checkers (e.g., edge detection for \texttt{rise}/\texttt{fall} expressions).
    \item \textbf{Method Registry:} To decouple the OSC2 language ontology from the CARLA API, a decorator-based \texttt{MethodRegistry} dynamically dispatches abstract AST actions (e.g., \texttt{vehicle.drive()}) to concrete, atomic Python behaviors (e.g., \texttt{WaypointFollower}, \texttt{ChangeTargetSpeed}). This registry centralizes the resolution of universal movement constraints, ensuring consistent application of collision avoidance and traffic rule adherence across all generated behaviors.
\end{itemize}

\textbf{Runtime Context Management:} Because certain parameters, such as relative speeds or dynamically calculated lane offsets, cannot be fully resolved at compile time, the backend relies on an \texttt{ExecutionContext}. Functioning as a lightweight Runtime State Manager, it recursively evaluates AST expression nodes during execution. By implementing $\mathcal{O}(1)$ caching mechanisms for live actor lookups and real-time physical unit conversions, declarative statements can be continuously re-evaluated as mathematical primitives within the simulation loop. 

Finally, before the primary Behavior Tree is ticked, a \texttt{ScenarioInitializer} parses the AST for constraints designated with the \texttt{at: start} modifier. It calculates spatial dependencies and executes the lazy-spawning or teleportation of actors, ensuring the concrete simulation state strictly aligns with the declarative initial conditions of the OSC2 script.
\begin{table*}[!h]
\centering
\small
\caption{Extended capability checklist aligned with OCS2, detailing supported actions, modifiers, and condition semantics within the implementation.}
\label{tab:osc_extended_checklist}
\begin{tabularx}{\textwidth}{@{} >{\raggedright\arraybackslash}p{0.15\textwidth} >{\raggedright\arraybackslash}p{0.15\textwidth} >{\raggedright\arraybackslash}X >{\raggedright\arraybackslash}X c @{}}
\toprule
\textbf{Category} & \textbf{Capability} & \textbf{Description} & \textbf{Implementation Method} & \textbf{Supported} \\
\midrule
\multirow{4}{=}{\textbf{Action Framework}}  
& Action lifecycle & Start, end, fail states & Behavior Tree status tracking & $\checkmark$ \\
& Actor binding & Map actions to actors & Runtime execution context & $\checkmark$ \\
& Composition & Sequential/parallel & Tree composites (serial/parallel) & $\checkmark$ \\
& Duration handling & Time bounds & Parallel timeout wrappers & $\checkmark$ \\
\midrule
\multirow{4}{=}{\textbf{Movement}}  
& Move/drive/walk & Motion primitives & LocalPlanner/WaypointFollower & $\checkmark$ \\
& Speed control & Adjust speed & PID longitudinal control & $\checkmark$ \\
& Acceleration control & Adjust acceleration & Direct physics/PID application & $\checkmark$ \\
& Stationary & Hold position & Zero-velocity kinematic lock & $\checkmark$ \\
\midrule
\multirow{4}{=}{\textbf{Position \& Path}}  
& Assign position & Set location & Absolute spatial transforms & $\checkmark$ \\
& Assign orientation & Set rotation & Rotational math application & $\checkmark$ \\
& Follow path & Move along path & Spline interpolation routing & $\checkmark$ \\
& Follow trajectory & Move along trajectory & Time-parameterized tracking & $\checkmark$ \\
\midrule
\multirow{3}{=}{\textbf{Interaction}}  
& Time gap & Time-based spacing & Dynamic setpoint interpolation & $\checkmark$ \\
& Space gap & Distance-based spacing & Topological route tracing & $\checkmark$ \\
& Headway & Relative positioning & Vector projections & $\checkmark$ \\
\midrule
\multirow{3}{=}{\textbf{Environment}}  
& Weather control & Change weather & Direct simulation API & $\checkmark$ \\
& Traffic signals & Control traffic lights & Semantic/Group state dispatch & $\checkmark$ \\
& Road conditions & Change surface/state & Friction trigger spawning & $\checkmark$ \\
\midrule
\multirow{8}{=}{\textbf{Modifiers}}  
& Speed modifier & Velocity profile & Dynamic profile parsing & $\checkmark$ \\
& Acceleration modifier & Acceleration profile & PID interpolation & $\checkmark$ \\
& Position modifier & Spatial constraints & Initialization placement context & $\checkmark$ \\
& Lateral modifier & Lane/side shift & OpenDRIVE lane offsets & $\checkmark$ \\
& Physical movement & Physics toggle & Simulation engine override & $\checkmark$ \\
& Temporal modifiers & Time limits/delays & Elapsed time evaluation & $\checkmark$ \\
& Relative modifiers & Relative offsets & Dynamic object referencing & $\checkmark$ \\
& Orientation modifiers & Rotation offsets & Yaw/Pitch/Roll application & $\checkmark$ \\
\midrule
\multirow{3}{=}{\textbf{Coordinate Systems}}  
& World coordinates & Global frame & Cartesian (x-y-z) transformations & $\checkmark$ \\
& Relative coordinates & Actor frame & Reference-based projection & $\checkmark$ \\
& Route-based (s-t) & Road frame (s-t) & OpenDRIVE coordinate mapping & $\checkmark$ \\
\midrule
\multirow{3}{=}{\textbf{Scenario Composition}}  
& Serial execution & Ordered steps & Sequence node construction & $\checkmark$ \\
& Parallel execution & Concurrent steps & Parallel node synchronization & $\checkmark$ \\
& Conditional triggers & Event-driven execution & Blackboard signal evaluation & $\checkmark$ \\
\midrule
\multirow{3}{=}{\textbf{Extensibility}}  
& Custom actions & Custom behavior & MethodRegistry decorators & $\checkmark$ \\
& Conflict resolution & Handle conflicts & Blackboard arbitration logic & $\checkmark$ \\
& Semantic validation & Check logic & AST resolution pass & $\checkmark$ \\
\midrule
\multirow{3}{=}{\textbf{Condition \& Expression}}  
& one\_of & Pick one value & \texttt{SuccessOnOne} tree policy & $\checkmark$ \\
& rise & False $\rightarrow$ True & Edge detection condition & $\checkmark$ \\
& fall & True $\rightarrow$ False & Edge detection condition & $\checkmark$ \\
\bottomrule
\end{tabularx}
\end{table*}

\section{Simulation Ontology \& Execution Mapping} 
\label{sec:ontology}

Executing a OSC2 scenario requires mapping the corresponding domain model/simulation ontology (\texttt{domain.osc}) to discrete simulation commands. This is achieved by binding the abstract ontology directly to the CARLA API via the intermediate atomic behaviors registered within the \texttt{MethodRegistry} \cite{dosovitskiy2017carla, asam2024openscenariodsl}. 

To define the exact scope of this integration, Table \ref{tab:osc_extended_checklist} provides a comprehensive checklist of the supported actions, modifiers, and condition semantics successfully implemented within the compiler framework.

\subsection{Execution of Actions and Modifiers}

To satisfy the declarative constraints of OSC2 DSL, core movement actions (e.g., \texttt{vehicle.drive()}, \texttt{person.walk()}) are systematically mapped to underlying atomic behaviors. The runtime evaluation engine dynamically delegates parameters to these behaviors based on the active modifiers present in the AST. 

As detailed in Table \ref{tab:osc_extended_checklist}, the compiler handles two primary categories of dynamic modifiers to execute these actions:
\begin{itemize}
    \item \textbf{Kinematic Modifiers:} Constraints such as \texttt{speed}, \texttt{change\_speed}, and \texttt{acceleration} calculate target scalar velocities or force profiles. These are executed via continuous dynamic adjustments using Proportional-Integral-Derivative (PID) control, allowing for relative evaluations (e.g., \texttt{faster\_than}) against other simulated entities.
    \item \textbf{Spatial Modifiers:} Constraints such as \texttt{position}, \texttt{lane}, \texttt{keep\_lane}, and \texttt{change\_lane} utilize OpenDRIVE standard logic and topological map projections. This enables the localized planner to calculate lateral offsets, dynamic splines, and target waypoints while evaluating collision avoidance and traffic light compliance.
\end{itemize}

This explicit mapping supports a comprehensive matrix of atomic behaviors, effectively transforming declarative OpenSCENARIO constraints into continuous, tick-by-tick waypoint tracking, velocity control, and environmental state loops.
\section{Case Study: Dynamic Cut-In and Evasive Maneuver} \label{sec:case_study}

To validate the expressiveness of the developed compiler architecture and its adherence to the standard, a comprehensive multi-actor test scenario\footnote{A video demonstration of this scenario execution is available at: \url{https://youtu.be/XrHTOlMSTpg}} is presented and evaluated herewith. In this particular case, the scenario specifies an adversarial interaction wherein a Heavy Goods Vehicle (HGV) rapidly overtakes and cuts in front of an \textit{ego} vehicle (the ``\textit{hero}''). The HGV subsequently performs a sudden deceleration \textit{(a.k.a. brake check)}, forcing the \textit{ego} vehicle to execute a concurrent evasive lane change and visual warning (flashing high beams). Finally, both vehicles synchronize their deceleration to stop safely before a static obstacle.

This scenario tests the runtime's ability to evaluate dynamic mathematical expressions, continuous spatial queries, and multi-actor event synchronization.

\subsection{Declarative Initialization and Topology}

The scenario script begins by defining the ontological entities, resolving dynamic state variables, and projecting the actors onto the topological road network.

\begin{lstlisting}[language=Python, numbers=left, caption={OSC2 Declarations, Global Variables, and Initialization Phase.}, label=lst:osc_init]
import "domain.osc"
use std.stdtypes

scenario hello_world:
  # --- DEFINITIONS ---
  carla_map: map with:
    keep(it.map_file == "Town06")
    
  env: environment
  
  hero: vehicle with:
    keep(it.model == "vehicle.tesla.model3")
    keep(it.name == "hero")

  npc: vehicle with:
    keep(it.model == "vehicle.carlamotors.european_hgv")
    keep(it.name == "npc")
    keep(it.color == "0,128,0")

  obstacle: stationary_object with:
    keep(it.name == "obstacle")
    keep(it.model == "static.prop.trafficwarning")

  # --- GLOBAL VARIABLES ---
  var v_hero: speed = 35kph
  var v_npc_fast: speed = v_hero + 12.42mph
  var v_npc_slow: speed = v_hero - 10kph      
  var v_npc_catchup: speed = v_hero * 10kph
    
  var lag: length = 5m              
  var gap: length = lag * 3              
  var safety_gap: length = gap - 3m      

  do serial:
    # --- TWILIGHT SETTINGS --- 
    env.assign_celestial_position(azimuth: 270deg, elevation: 12deg)
    hero.set_lights(mode: "auto") 

    hero.assign_position() with:
      lane(1, at: start)
      speed(0kph, at: start)

    npc.assign_position() with:
      lane(side: right, side_of: hero, at: start)
      position(distance: lag, behind: hero, at: start)
      speed(0kph, at: start)

    obstacle.assign_position() with:
      position(x: 478.93, y: -14.07, z: 0.00, h: -1.57rad, at: start)
\end{lstlisting}

As shown in Listing \ref{lst:osc_init}, the \texttt{ExecutionContext} processes relative assignments (e.g., \texttt{v\_npc\_fast = v\_hero + 12.42mph}) prior to execution. Leveraging \texttt{stdtypes} definitions, the compiler recursively evaluates and converts mixed dimensional quantities (e.g., mph, kph) into CARLA's native SI metric units (m/s, m) before passing values to the simulation engine.

During initialization (\texttt{at: start}), the \texttt{ScenarioInitializer} executes dynamic actor placement across three instantiation paradigms:
\begin{itemize}
    \item \textbf{Default Map:} Generic constraints (e.g., \texttt{hero}) query the CARLA API for predefined valid spawn points, ensuring safe default placement \textit{(lines 39--41)}.
    \item \textbf{Relative Topological:} The \texttt{npc} is spawned relative to \texttt{hero} by resolving the OpenDRIVE spline, applying lateral lane offsets, and projecting backward by the \texttt{lag} distance \textit{(lines 43--46)}.
    \item \textbf{Absolute Cartesian:} Static props like the \texttt{obstacle} bypass topological tracing, using explicit world coordinates \textit{(lines 48--49)}.
\end{itemize}

\begin{figure}[!hb]
    \centering
    \begin{subfigure}{\linewidth}
    \centering
        \includegraphics[width=.85\linewidth]{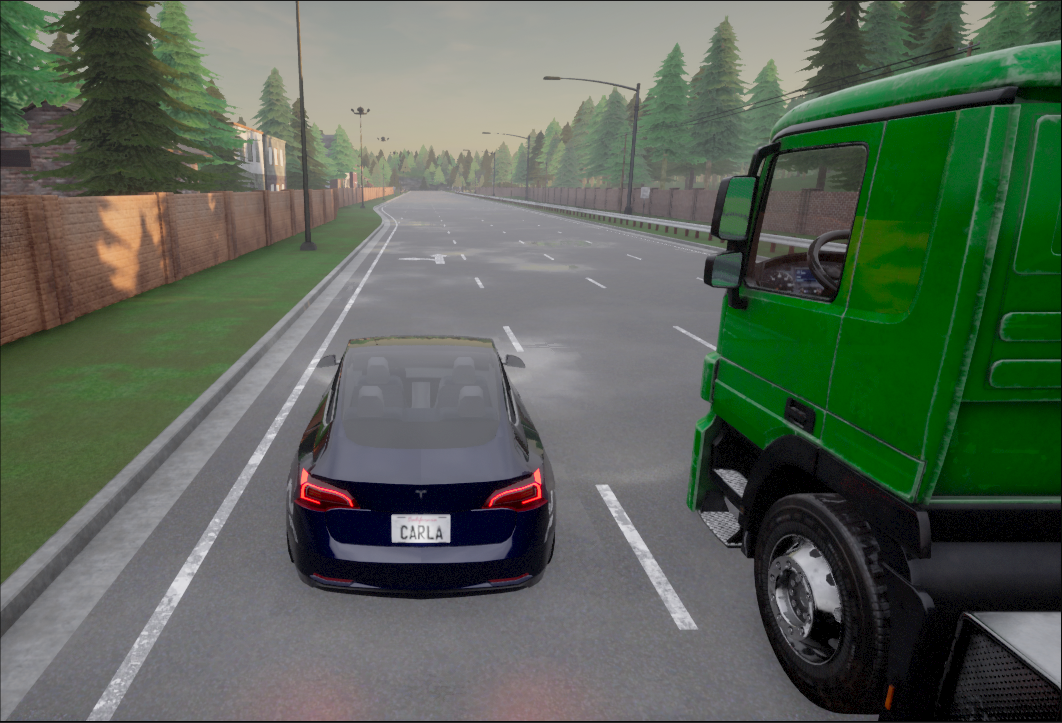}
        \caption{Default spectator view depicting the relative longitudinal and lateral offsets.}
        \label{fig:init-1}
    \end{subfigure}
    
    \vspace{0.5em}
    
    \begin{subfigure}{\linewidth}
    \centering
        \includegraphics[width=.85\linewidth]{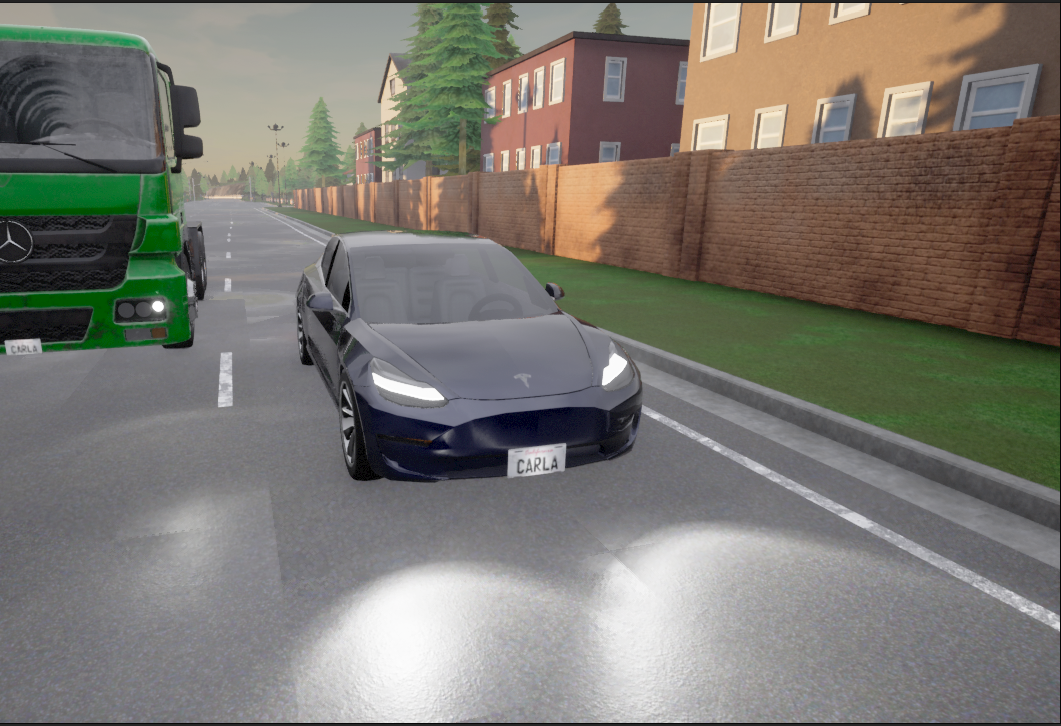}
        \caption{Frontal perspective illustrating the twilight environmental conditions and automatically engaged daytime running lights.}
        \label{fig:init-2}
    \end{subfigure}
    \caption{Simulation state following the declarative initialization phase, captured from multiple perspectives.}
    \label{fig:scenario_init}
\end{figure}

To reliably track entities, the compiler binds the OSC2 \texttt{keep(it.name == "...")} constraint to CARLA's \texttt{role\_name} attribute. This deterministic mapping is critical for ROS 2 integration, enabling the \texttt{Ros2SubscriberManager} to automatically configure sensor namespaces (e.g., \texttt{/carla/hero/imu\_sensor}) and control topics without manual overhead.

Furthermore, the \texttt{environment} pseudo-actor declaratively controls the weather. The \texttt{assign\_celestial\_position} action translates azimuth and elevation into CARLA sun angles. As altitude drops, the \texttt{MethodRegistry} detects the shift and automatically activates the ego vehicle's dynamic lighting, linking global conditions to atomic behaviors. Crucially, this \texttt{env} declaration scales beyond initialization, allowing dynamic mid-simulation injections of precipitation, fog, or time-of-day progression for multi-phase meteorological testing.

Figure \ref{fig:scenario_init} illustrates the resulting state, highlighting accurate relative placement and twilight environmental parameters.

\subsection{Ego Vehicle (Hero) Execution Logic}
Following initialization, the runtime transitions to the main execution blocks. The ego vehicle's behavior tree demonstrates continuous state evaluation and parallel execution capabilities.
\newpage
\begin{lstlisting}[language=Python, numbers=left, firstnumber=last, caption={Ego Vehicle (Hero) behavior logic detailing continuous queries and parallel execution.}, label=lst:osc_hero]
  do parallel:
    # ==========================
    # --- HERO LOGIC ---
    # ==========================
    serial:
      wait @go_signal

      # --- Phase 1: Cruise ---
      one_of:
        hero.drive() with:
          speed(v_hero)
        wait fall(npc.position.ahead_of(hero) > safety_gap)

      # --- Phase 2: Hazard Detected - Flash High Beams & Swerve ---
      parallel:
        # Task A: The evasive physical maneuver
        serial:
          hero.change_lane(num_of_lanes: 1, side: right)
          emit CRASH_AVOIDED
                
        # Task B: The visual "Flash" effect
        serial:
          hero.set_lights(mode: "high_beam")
          wait elapsed(0.5s)
          hero.set_lights(mode: "auto")

      # --- Phase 3: Approach & Synchronize ---
      one_of:
        hero.drive() with:
          speed(v_hero)
        wait @OBSTACLE_DETECTED                

      # --- Phase 4: Approach NPC smoothly and pull alongside ---
      serial:
        hero.change_speed(target: 25kph, rate_profile: smooth)
        one_of:
          hero.drive() with:
            speed(25kph)
          serial:
            wait rise(hero.position.ahead_of(npc) >= -1m)

      # --- Phase 5: Final Stop ---
      hero.change_speed(target: 0kph, rate_profile: asap)
      wait hero.speed < 0.1kph
      wait elapsed(5s)
\end{lstlisting}

Listing \ref{lst:osc_hero} highlights the Condition and Expression layer. In Phase 1, the \texttt{hero} executes \texttt{drive()} within a \texttt{one\_of} composite alongside a \texttt{wait} directive. The \texttt{fall(...)} argument forces the runtime to continuously query topological network distances, triggering an edge detector that terminates the drive action and transitions to Phase 2 when the threshold is crossed.

\begin{figure}[!ht]
    \centering
    \begin{subfigure}{\linewidth}
    \centering
        \includegraphics[width=.85\linewidth]{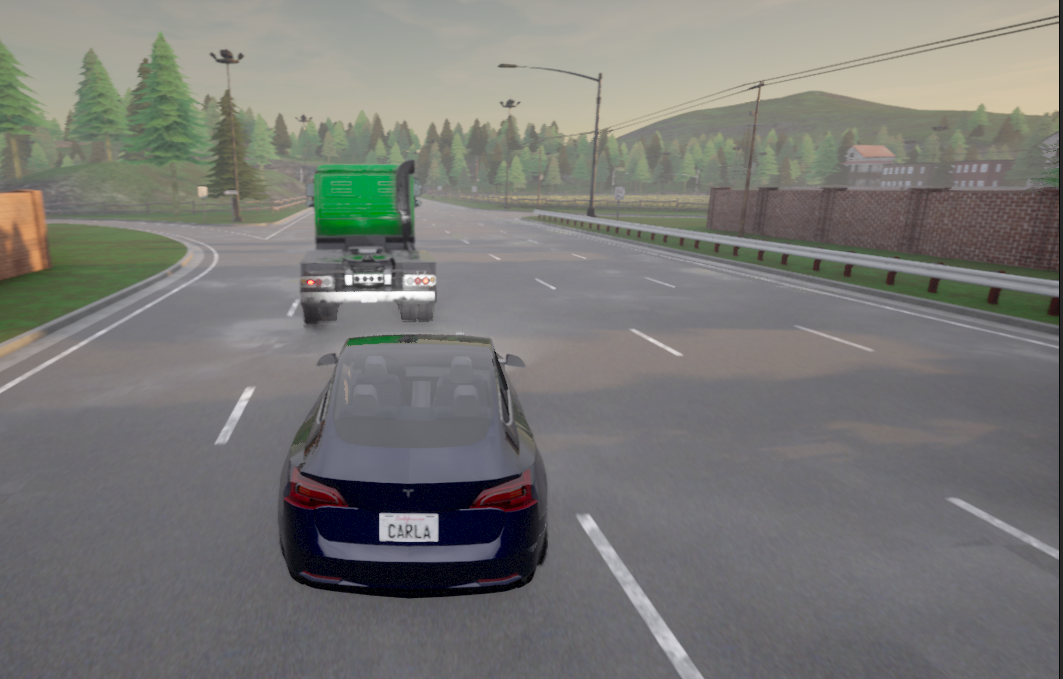}
        \caption{Exterior perspective capturing the exact moment the ego vehicle initiates the evasive right lane change and activates its high beams.}
        \label{fig:lane-change-1}
    \end{subfigure}
    
    \vspace{0.5em}
    
    \begin{subfigure}{\linewidth}
    \centering
        \includegraphics[width=.85\linewidth]{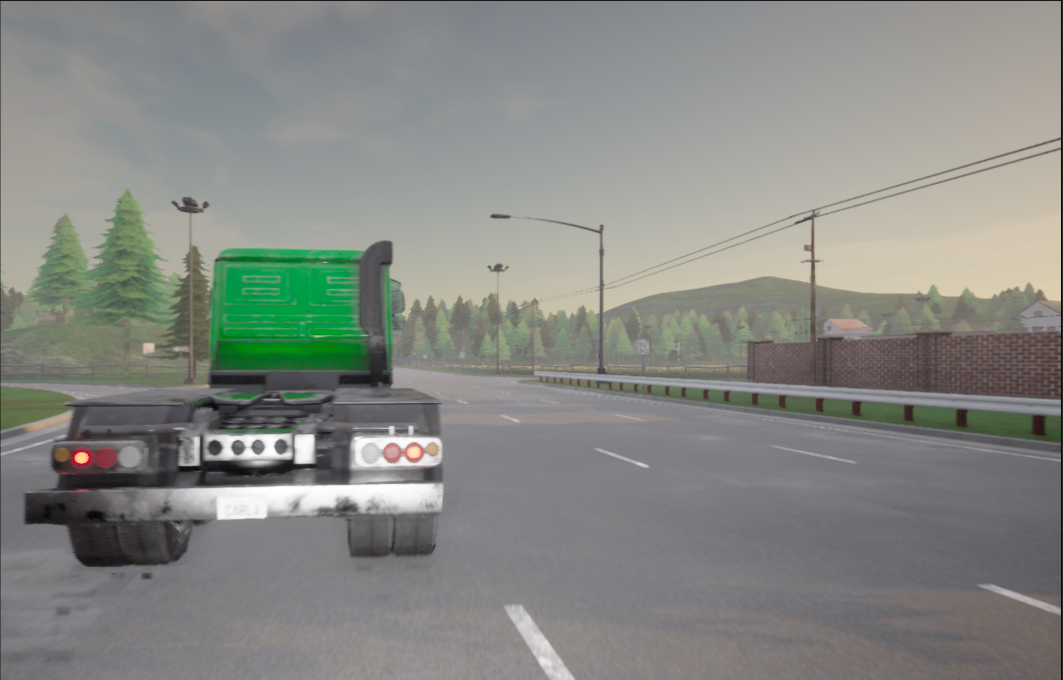}
        \caption{Driver's point-of-view (POV) at the identical simulation tick, illustrating the visual warning effect from within the cabin.}
        \label{fig:lane-change-2}
    \end{subfigure}
    \caption{Phase 2 execution: The ego vehicle concurrently executing a lateral maneuver and a visual warning in response to the spatial trigger.}
    \label{fig:scenario_evasive}
\end{figure}

Phase 2 demonstrates the \texttt{parallel} composite. The \texttt{BehaviorTreeBuilder} generates a concurrent node combining a physical \texttt{change\_lane} maneuver with a simultaneous vehicle light state update (flashing high beams). An internal \texttt{CRASH\_AVOIDED} event is emitted as a synchronization flag for the adversarial actor. Figure \ref{fig:scenario_evasive} captures this exact tick, showing both the exterior lateral trajectory and lighting (Figure \ref{fig:lane-change-1}) alongside the interior visual feedback (Figure \ref{fig:lane-change-2}).

Phase 3 highlights decoupled synchronization via a blackboard architecture. The \texttt{wait @OBSTACLE\_DETECTED} directive suspends the sequence until a global flag is raised, permitting the \texttt{hero} to cruise indefinitely until an external trigger broadcasts the signal. This facilitates complex inter-actor orchestration without hardcoded temporal dependencies.

Phase 4 illustrates advanced kinematic control. The \texttt{MethodRegistry} maps the \texttt{rate\_profile: smooth} modifier to a highly damped PID controller within the \texttt{ChangeTargetSpeed} atomic behavior, ensuring realistic, jerk-limited deceleration. Concurrently, a rising edge condition (\texttt{wait rise(...)}) monitors relative spatial queries dynamically to detect when the ego vehicle pulls alongside the \texttt{npc}.

Phase 5 concludes with an \texttt{asap} speed profile, bypassing smoothed PID gains for maximum braking. The subsequent \texttt{wait hero.speed < 0.1kph} directive demonstrates direct member access: the \texttt{ExecutionContext} fetches the vehicle's 3D velocity vector per-tick, computes its scalar magnitude, and evaluates the relation. The sequence terminates cleanly via a non-blocking timeout composite generated by \texttt{wait elapsed(5s)}.

\subsection{Adversarial (NPC) Execution Logic}

The adversarial logic executes concurrently with the ego vehicle, relying on dynamic speed interpolation and cross-actor event synchronization.

\begin{lstlisting}[language=Python, numbers=left, firstnumber=last, caption={NPC behavior logic demonstrating event synchronization and Euclidean distance queries.}, label=lst:osc_npc]
  # ==========================
  # --- NPC LOGIC ---
  # ==========================
  serial:
    wait @go_signal
        
    # --- Phase 1: Accelerate UNTIL ahead of Hero ---
    one_of:
      npc.drive() with:
        speed(v_npc_fast)
      serial:
        wait rise(npc.position.ahead_of(hero) >= lag * 2)

    # --- Phase 2: Cut In ---
    npc.change_lane(num_of_lanes: 1, side: left)

    # --- Phase 3: Brake Check UNTIL Ego swerves ---
    one_of:
      npc.drive() with:
          speed(v_npc_slow)
      wait @CRASH_AVOIDED

    # --- Phase 4: Recover & Approach Obstacle ---
    one_of:
      npc.drive() with:
          speed(v_npc_catchup, rate_profile: smooth)
      serial:
          wait rise(npc.object_distance(reference: obstacle, direction: euclidean) < 45m)

    # --- Phase 5: Emergency Stop & Emit Event ---
    npc.change_speed(target: 0kph, rate_profile: asap)
    emit OBSTACLE_DETECTED
                    
    # Phase 6: Parked
    wait elapsed(100s)
\end{lstlisting}

\begin{figure}[ht]
    \centering
    \begin{subfigure}{\linewidth}
    \centering
        \includegraphics[width=.85\linewidth]{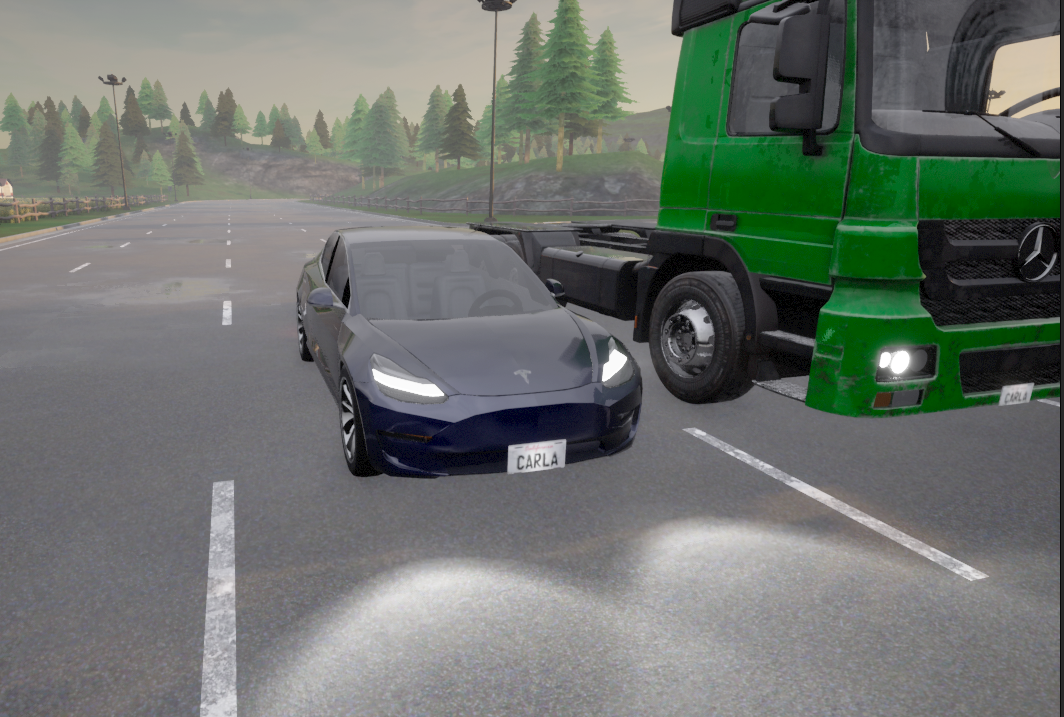}
        \caption{Frontal perspective capturing the synchronized halt of the ego vehicle and the HGV before the static obstacle.}
        \label{fig:end-1}
    \end{subfigure}
    
    \vspace{0.5em}
    
    \begin{subfigure}{\linewidth}
    \centering
        \includegraphics[width=.85\linewidth]{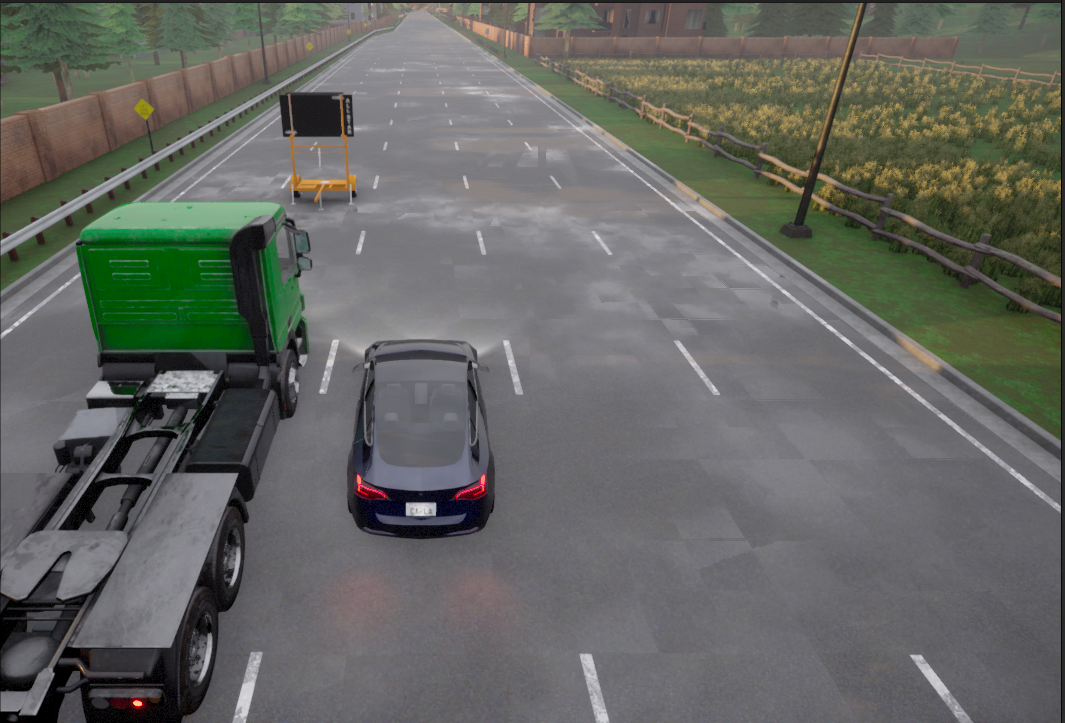}
        \caption{Bird's-eye view confirming the final lateral alignment and longitudinal gap maintenance at the scenario's conclusion.}
        \label{fig:end-2}
    \end{subfigure}
    \caption{Phase 5 execution: Both actors synchronize their deceleration profiles to halt safely, demonstrating precise spatial and temporal control.}
    \label{fig:scenario_stop}
\end{figure}

Listing \ref{lst:osc_npc} details the adversarial \texttt{npc} logic. Building upon the constructs introduced in the ego vehicle's routine, it highlights dynamic expression evaluation, bidirectional synchronization, and versatile spatial queries.

In Phase 1, the \texttt{one\_of} composite evaluates a topological distance query against a dynamically computed threshold (\texttt{lag * 2}). This demonstrates the \texttt{ExecutionContext}'s ability to resolve mathematical expressions continuously at runtime, bypassing the need for static literals. Phase 2 then executes a standard lateral cut-in maneuver.

Phases 3 and 5 showcase asynchronous, cross-actor orchestration. The \texttt{npc} maintains a deliberate deceleration until receiving the \texttt{CRASH\_AVOIDED} signal from the ego vehicle's evasive maneuver. Conversely, the \texttt{npc}'s final emergency stop emits the \texttt{OBSTACLE\_DETECTED} event to release the ego vehicle from its suspended cruise. This bidirectional event handshaking achieves complex synchronization without rigid temporal scripting.

Phase 4 introduces a critical distinction in spatial resolution. While re-applying the previously discussed smoothed PID profile, it executes an \texttt{object\_distance} query with the \texttt{direction: euclidean} parameter. Unlike topological \texttt{ahead\_of} queries that trace OpenDRIVE road splines, this instructs the \texttt{MethodRegistry} to compute a direct 3D Cartesian distance. This mechanism is essential for detecting static props, such as the \texttt{obstacle}, which reside outside the bounds of the routable road network. Finally, Phase 6 utilizes a basic \texttt{elapsed} directive to indefinitely suspend the actor's execution tree.

\section{Discussion and Future Work}
\label{sec:discussion}

This paper presented a multi-pass compiler architecture mapping the declarative OpenSCENARIO v2.1 DSL to the procedural CARLA simulator. By decoupling parsing and semantic analysis from physics execution, the framework translates abstract intent into deterministic behavior trees. The dynamic runtime -- comprising the \texttt{ExecutionContext} and \texttt{MethodRegistry} -- enables real-time mathematical evaluation, continuous spatial queries, and multi-actor synchronization without external solvers or static bytecode.

Despite the baseline established, the architecture presents distinct limitations. First, while core kinematic and spatial modifiers are integrated, the expansive OSC2 ontology requires further expansion to cover complex domains like pedestrian intent, intersection right-of-way, and probabilistic weather. Second, continuously evaluating topological and Euclidean queries within a Python-based loop introduces computational overhead, potentially degrading real-time fidelity in high-density traffic scenarios.

Future work will prioritize architectural optimization and further standard compliance. To mitigate latency, expensive spatial queries will be migrated to lower-level C++ CARLA bindings. Additionally, \texttt{MethodRegistry} will be expanded to achieve complete ontological coverage and adapt the semantic analyzer for the upcoming OpenSCENARIO v2.2.0 syntax, ultimately facilitating massive-scale, reproducible scenario-based testing.
{
    \small
    \bibliographystyle{ieeenat_fullname}
    \bibliography{main}
}


\end{document}